\PassOptionsToPackage{table}{xcolor}
\documentclass[10pt,twocolumn,letterpaper]{article}
\usepackage[final]{cvpr} 
\usepackage{graphicx}
\usepackage{mathtools} 
\usepackage{graphicx}
\usepackage{amsmath}
\usepackage{amssymb}
\usepackage{booktabs}
\usepackage{dsfont}
\usepackage{multirow}
\usepackage{multicol}
\usepackage{makecell}
\usepackage[pagebackref,breaklinks,colorlinks]{hyperref}
\usepackage[usenames,dvipsnames]{xcolor}
\usepackage{tikz}
\usepackage{pgfplots}
\usepackage{pgfkeys}

\usepackage[font=small,labelsep=period]{caption}

\def\addlegendimage{\csname pgfplots@addlegendimage\endcsname}

\usepackage[capitalize]{cleveref}
\crefname{section}{Sec.}{Secs.}
\Crefname{section}{Section}{Sections}
\Crefname{table}{Table}{Tables}
\crefname{table}{Tab.}{Tabs.}

\usepackage{amsmath,amsfonts,bm}
\usepackage{pifont}

\def\eqref#1{equation~\ref{#1}}

\def\1{\bm{1}}

\DeclareMathAlphabet{\mathsfit}{\encodingdefault}{\sfdefault}{m}{sl}
\SetMathAlphabet{\mathsfit}{bold}{\encodingdefault}{\sfdefault}{bx}{n}

\newcommand{\NumTestObjects}[0]{\text{N}_\text{o}}
\newcommand{\NumProposals}[0]{\text{N}_\text{p}}
\newcommand{\NumViews}[0]{\text{V}}
\newcommand{\DescriptorSize}[0]{\text{C}}

\newcommand{\Mask}[0]{\text{M}}
\newcommand{\ObjectID}[0]{\text{o}}
\newcommand{\Score}[0]{\text{s}}

\newcommand{\ReferenceDescriptor}[0]{\mathbf{D}_\textbf{r}}
\newcommand{\ProposalDescriptor}[0]{\mathbf{D}_\textbf{p}}

\def\cvprPaperID{9} 
\def\confName{ICCV}
\def\confYear{2023}
\title{CNOS: A Strong Baseline for CAD-based Novel Object Segmentation

}

\newcommand{\namesep}{\hspace{0.8em}}
\author{
Van Nguyen Nguyen$^{1}$\namesep
Thibault Groueix$^{2}$\namesep
Georgy Ponimatkin$^{1}$\namesep
Vincent Lepetit$^{1}$\namesep
Tomas Hodan$^{3}$\\
{$^{1}$LIGM, \'Ecole des Ponts}\namesep{$^{2}$Adobe}\namesep{$^{3}$Reality Labs at Meta}\\[0.2cm]
Source code: \href{https://github.com/nv-nguyen/cnos}{https://github.com/nv-nguyen/cnos}\\
}

\begin{document}

\definecolor{darkgreen}{RGB}{0,110,0}
\definecolor{darkred}{RGB}{170,0,0}
\def\greencheckmark{\textcolor{darkgreen}{\checkmark}}
\def\redxmark{\textcolor{darkred}{\text{\ding{55}}}}  %

\definecolor{DarkMagenta}{rgb}{0.7, 0.0, 0.7}
\newcommand{\nguyen}[1]{{\color{DarkMagenta}#1}}
\newcommand{\nguyenrmk}[1]{{\color{DarkMagenta} {\bf [VN: #1]}}}

\definecolor{DarkBlue}{rgb}{0.0, 0.0, 0.8}
\newcommand{\thibault}[1]{{\color{DarkBlue} #1}}
\newcommand{\thibaultrmk}[1]{{\color{DarkBlue} {\bf [TG: #1]}}}

\definecolor{DarkOrange}{rgb}{1.0, 0.55, 0.0}
\newcommand{\tom}[1]{{\color{DarkOrange}#1}}
\newcommand{\tomrmk}[1]{{\color{DarkOrange} {\bf [TH: #1]}}}

\definecolor{DarkGreen}{rgb}{0.0, 0.5, 0.0}
\newcommand{\vincent}[1]{{\color{DarkGreen} #1}}
\newcommand{\vincentrmk}[1]{{\color{DarkGreen} {\bf [VL: #1]}}}

\newcommand{\was}[1]{}

\definecolor{Random}{rgb}{0.1, 0.5, 0.9}
\newcommand{\georgy}[1]{{\color{Random} #1}}
\newcommand{\georgyrmk}[1]{{\color{Random} {\bf [GP: #1]}}}

\newcommand\customparagraph[1]{\vspace{0.7em}\noindent\textbf{#1}}

\renewcommand{\nguyen}[1]{#1}
\renewcommand{\nguyenrmk}[1]{}
\renewcommand{\thibault}[1]{#1}
\renewcommand{\thibaultrmk}[1]{}
\renewcommand{\georgy}[1]{#1}
\renewcommand{\georgyrmk}[1]{#1}
\renewcommand{\tom}[1]{#1}
\renewcommand{\tomrmk}[1]{}
\renewcommand{\vincent}[1]{#1}
\renewcommand{\vincentrmk}[1]{}

\maketitle

\begin{abstract}
We propose a simple yet powerful method to segment novel objects in RGB images from their CAD models.
Leveraging recent foundation models, Segment Anything and DINOv2, we generate segmentation proposals in the input image and match them against object templates that are pre-rendered using the CAD models. The matching is realized by comparing DINOv2 \texttt{cls} tokens of the proposed regions and the templates. The output of the method is a set of segmentation masks associated with per-object confidences defined by the matching scores.
We experimentally demonstrate that the proposed method achieves state-of-the-art results in CAD-based novel object segmentation on the seven core datasets of the BOP challenge, surpassing the recent method of Chen~\textit{et al.} by absolute 19.8\% AP. 
\end{abstract}

\thispagestyle{plain}
\pagestyle{plain}


\section{Introduction}

\label{sec:introduction}

Object pose estimation plays a critical role in robotics and augmented reality applications. While supervised deep learning methods have achieved remarkable performance, they rely on extensive training data specific to each target object~\cite{labbe-eccv20-cosypose,Wang_2021_GDRN, sundermeyer2023bop}. Introducing objects unseen during training therefore requires a significant effort to synthesize or annotate data and retrain the model. This restricts the application of the supervised methods in industry. For instance, in a logistic warehouse, it appears impractical to retrain the pose estimation method for every new product.

Performing object pose estimation typically involves two main steps: (1) the target objects are detected/segmented in the input image, and (2) the 6D object poses are then estimated from the detected regions~\cite{sundermeyer2023bop}. Recent works such as template-pose~\cite{nguyen2022templates} and MegaPose~\cite{megapose} introduced effective CAD-based object pose estimation methods. However, these methods mainly focus on the second step and require input 2D bounding boxes, which restricts their applicability to scenarios where precise 2D bounding boxes are available.

\begin{figure}
\centering
    
\begin{tikzpicture}
    \tikzstyle{every node}=[font=\small]
    \begin{axis}[
        width=7.5cm,
        height=5.5cm,
        font=\footnotesize,
        xlabel=Run-time (s),
        ylabel=Average Precision ($\%$),
        ymin=0,
        ymax=80,
        ytick={20, 40, 60, 80},
        xmin=0,
        xmax=1.2,
        xtick={0, 0.25, 0.5, 0.75, 1, 1.25},
        label style={font=\small},
        tick label style={font=\small},
        legend pos=north east,
        grid=major,
        legend style={nodes={scale=0.9, transform shape}},
        legend cell align={left}
    ]

    \addlegendimage{only marks, mark=x, color=blue, mark size=4pt}
    \addlegendentry{Supervised}
    
    \addlegendimage{only marks, mark=*, color=Orange, mark size=3pt}
    \addlegendentry{Unsupervised}
    
    \draw plot[mark=x, color=blue, mark size=2pt,mark options={color=blue,mark size=4pt}] coordinates {(axis cs: 0.054,39.2)};
    \node[anchor=west] at (axis cs:0.0,31.2) {\small Mask R-CNN \cite{he2017mask}};
    
    \draw plot[mark=x, color=blue, mark size=2pt,mark options={color=blue,mark size=4pt}] coordinates {(axis cs: 0.08,57.8)};
    \node[anchor=west] at (axis cs:0.00,63.8) {ZebraPose \cite{zebrapose}};

    \draw plot[mark=*, color=Orange, mark size=5pt,mark options={color=Orange,mark size=2pt}] coordinates {(axis cs: 1,21.4)};
    \node[anchor=west] at (axis cs:0.8,26.4) {Chen~\textit{et al.}~\cite{chen20233d}};

    \draw plot[mark=*, color=Orange, mark size=3pt,mark options={color=Orange,mark size=2pt}] coordinates {(axis cs: 0.34,41.2)};
    \node[anchor=west, font=\small] at (axis cs:0.21,46.2) {\small CNOS (Ours)};

    \end{axis}
\end{tikzpicture}
\vspace{-0.1cm}
 \caption{\textbf{Performance on the seven core BOP datasets~\cite{sundermeyer2023bop}}. Our method, CNOS, relies on FastSAM~\cite{zhao2023fast} for generation of segmentation proposals and on DINOv2~\cite{oquab2023dinov2} for visual description. CNOS outperforms the unsupervised method of Chen \textit{et al.}~\cite{chen20233d} and even the supervised method Mask R-CNN~\cite{he2017mask}, which was trained on tens of thousands of images per BOP dataset and used in CosyPose~\cite{labbe-eccv20-cosypose}. Similarly to Mask R-CNN~\cite{he2017mask}, the runtime of CNOS is dominated by the proposal stage.}
\label{fig:teaser}
\end{figure}
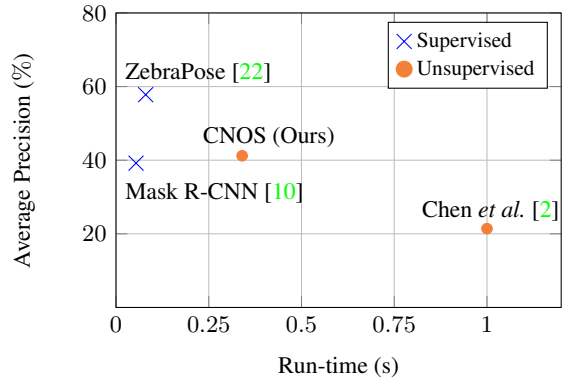

To address the gap, we propose a simple method for object detection and segmentation that only requires CAD models of the target objects. The method is dubbed {\bf CNOS} for {\bf C}AD-based {\bf N}ovel {\bf O}bject {\bf S}egmentation.

In CNOS, new objects are onboarded by rendering their CAD models and describing each rendered template by the DINOv2 \texttt{cls} token~\cite{oquab2023dinov2}. Given an RGB input image, segmentation proposals are extracted from the image by Segment Anything~(SAM)~\cite{kirillov2023segment} or Fast Segment Anything~(FastSAM)~\cite{zhao2023fast} and matched against the templates based on the similarity between their DINOv2 \texttt{cls} tokens. Rendering the templates takes less than 2 seconds per CAD model which is much faster than retraining of supervised methods, which typically requires several hours.
The choice of DINOv2 for measuring the similarity between templates and proposals is mainly motivated by its ability to effectively address the domain gap between real and synthetic images. We also demonstrate that photo-realistic rendering techniques of BlenderProc \cite{denninger2019blenderproc}, which require approximately 1 second to render an image, can be leveraged to further mitigate this domain gap and enhance accuracy.
Experiments on the seven core datasets of the BOP challenge~\cite{sundermeyer2023bop}
demonstrate the state-of-the-art performance of CNOS. 

As shown in Figure~\ref{fig:teaser}, CNOS outperforms the recent unsupervised method for CAD-based segmentation by Chen \textit{et al.}~\cite{chen20233d} 
and even Mask R-CNN \cite{he2017mask}, a supervised method that was trained on tens of thousands of images per BOP dataset and used in CosyPose~\cite{labbe-eccv20-cosypose}.

\begin{figure*}[t]
    \centering
    \includegraphics[width=0.95\textwidth]{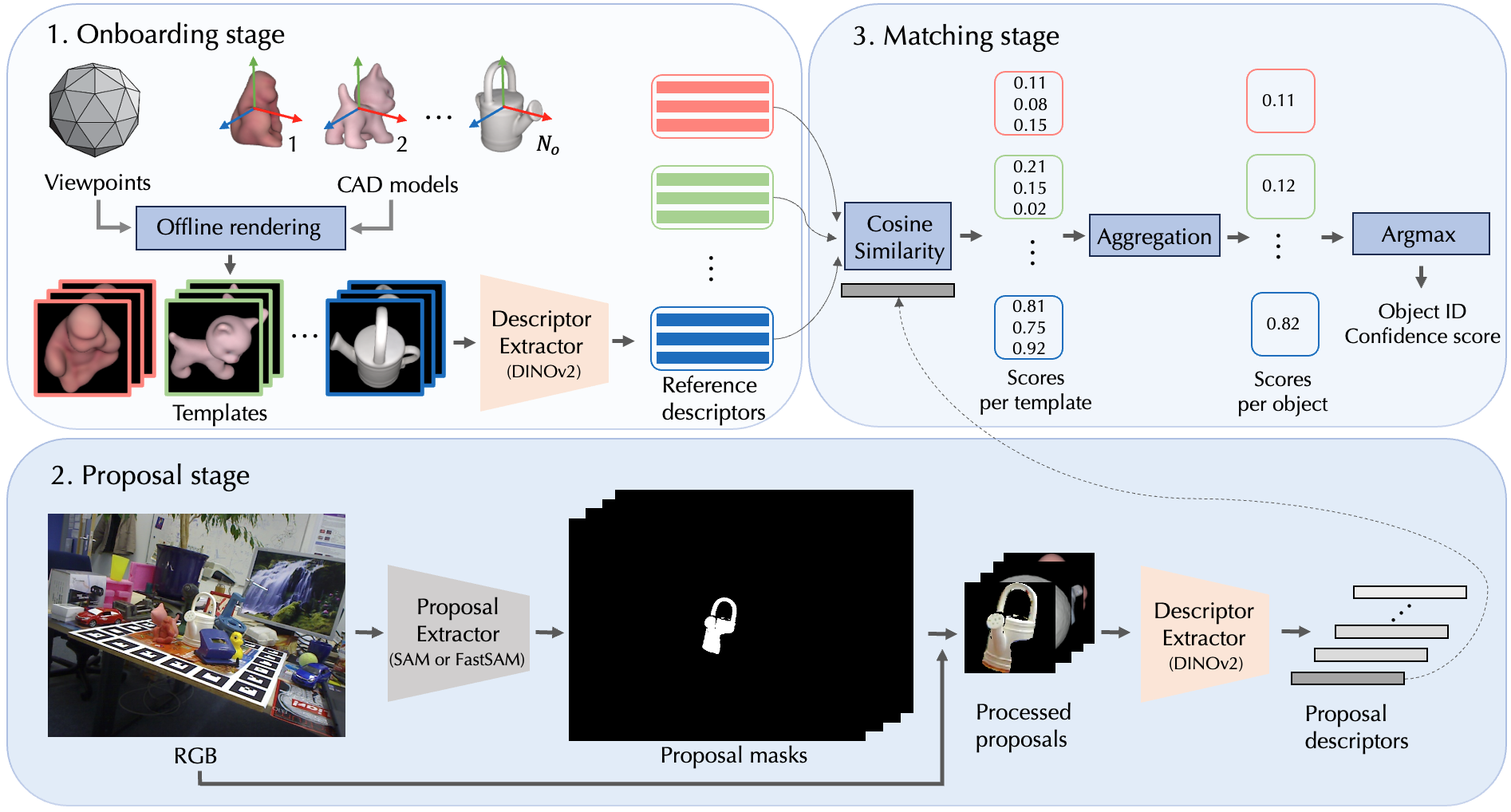}
    \caption{{\bf CNOS overview.} 
    Given CAD models of the target objects, the objects are onboarded by (i) rendering a set of templates showing the models from different viewpoints, and (ii) describing the templates by the DINOv2 \texttt{cls} token~(Section~\ref{sec:onboarding}). 
    At inference time, segmentation proposals are generated from the input RGB image using SAM or FastSAM~(Section \ref{sec:proposal}), and the proposals are matched against the templates by comparing their DINOv2 \texttt{cls} tokens~(Section~\ref{sec:matching}).  
    }
    \label{fig:main figure}
\end{figure*}


\section{Related work}
\label{sec:relatedwork}


This section provides a brief overview of existing methods for object detection and segmentation that are commonly used in 6D object pose estimation pipelines.

\customparagraph{Segmentation of seen object.} 
Many object pose estimation methods~\cite{labbe-eccv20-cosypose,Wang_2021_GDRN} employ object segmentation methods such as Mask R-CNN~\cite{he2017mask}, typically fine-tuned on extensive training data specific to each target object \cite{sundermeyer2023bop}.
Such supervised methods have been demonstrated robust in challenging scenarios with heavy occlusions and lighting variations. However, these methods cannot deal with new objects without retraining, which is a deal-breaker for many applications. In this work, we address this limitation by focusing on the segmentation of previously-unseen objects from their CAD models without retraining.

\customparagraph{Segmentation of unseen objects.}
Object segmentation methods traditionally focus on scenarios known as ``closed-world" settings, where the training and test sets share the same object classes. Nevertheless, recent observations by Du~\textit{et al.}~\cite{du2021learning, Du_ICCVW21} suggest that class-agnostic instance segmentation networks can effectively generalize to previously unseen object classes. Building upon this insight, Zhao~\textit{et al.}~\cite{zhao2022ncdss} leverage saliency detection models to solve the novel class discovery task in 2D segmentation.
Nguyen~\textit{et al.}~\cite{nguyen_pizza_2022} propose a two-stage 6D tracking approach based on these observations. Their approach assumes the availability of an initial bounding box to segment object using \cite{Du_ICCVW21} and then propagates the box to next frames using optical flow. Subsequently, 6D tracking of novel objects is performed based on predicted object masks. In contrast, our objective solely focuses on segmenting objects in images derived from CAD models without initial boxes.

Commonly used in robotics, UOIS-Net~\cite{xie2021unseen} employs a two-stage approach to segment novel objects. It operates on the depth channel of captured RGB-D images to generate object instance center votes and assembles them into rough initial masks. These masks are subsequently refined using the RGB channels. Xiang~\textit{et al.}~\cite{xiang2021learning} also propose an RGB-D based method that uses learned feature embeddings and applies a mean shift clustering algorithm to discover and segment unseen objects. To avoid using depths, Durner et al.~\cite{durner_unknown_2021} use horizontal correlation to extract disparity RGB-based features and segment novel objects from stereo RGB images. It is worth noting that UOIS-Net~\cite{xie2021unseen}, Xiang et al.~\cite{xiang2021learning}, and Durner et al.~\cite{durner_unknown_2021} are RGB-D or stereo RGB approaches, while our method targets the segmentation of unseen objects from only a single RGB image and CAD models, which is more applicable.

Recently, Segment Anything (SAM)~\cite{kirillov2023segment} has introduced a powerful foundation model for image segmentation capable of segmenting all objects in a given RGB image. Chen~\textit{et al.}~\cite{chen20233d} utilize SAM to extract object proposals, which are then combined with visual clues extracted by ImageBlind~\cite{girdhar2023imagebind}. Feature matching is subsequently applied for CAD-based novel object segmentation. Experiments presented in this paper show that CNOS surpasses the method of Chen~\textit{et al.} by absolute 19.8\% AP.

\section{Method}
\label{sec:method}
In this section, we provide a detailed description of our three-stage approach for CAD-based novel object segmentation. We first describe the onboarding stage in Section~\ref{sec:onboarding}, where we extract visual descriptors from renderings of the CAD models. In Section~\ref{sec:proposal}, we explain the proposal stage, which involves obtaining all possible masks and their descriptors. Finally, in Section~\ref{sec:matching}, we discuss the matching stage, where object masks are retrieved and labeled based on visual descriptors of their CAD models.

\subsection{Onboarding stage}
\label{sec:onboarding}

\begin{figure}[t]
    \centering
    \includegraphics[width=0.9\columnwidth]{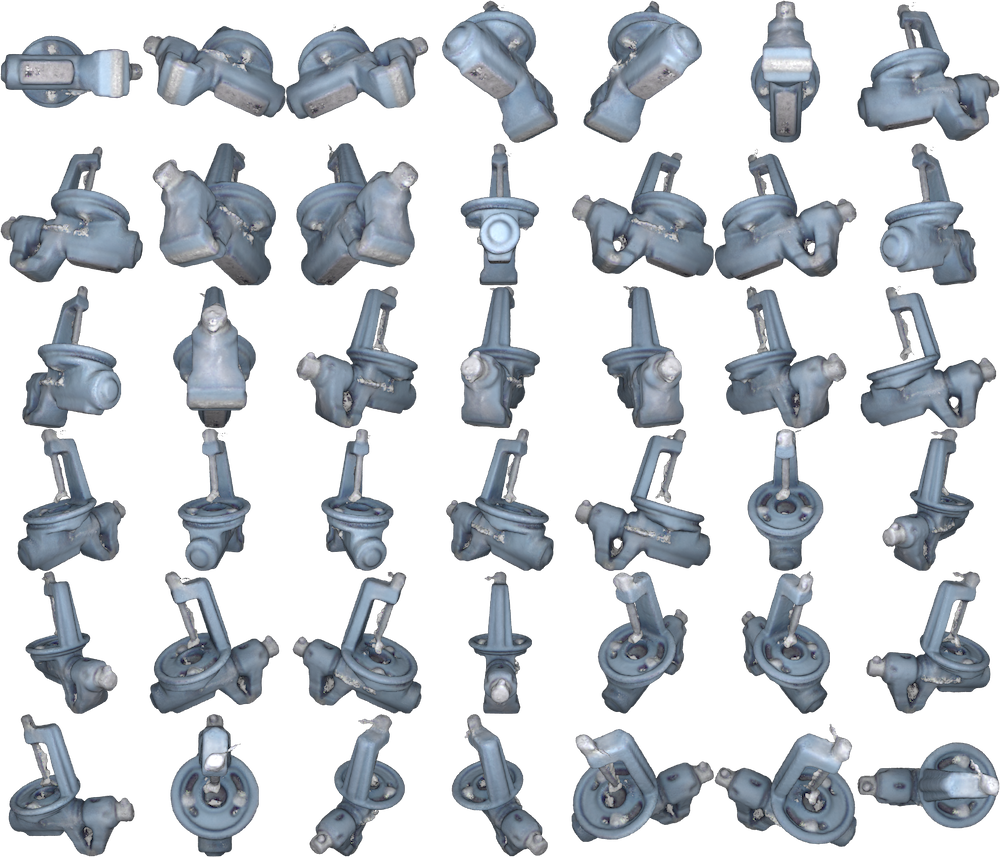}

    \vspace{2mm}
    \caption{{\bf Visualization of templates for the ``benchwise" object from LM-O \cite{hinterstoisser-accv12-modelbasedtrainingdetection} rendered with Pyrender \cite{Pyrender}.} 42 templates were rendered from viewpoints defined by the icosphere~\cite{nguyen2022templates}.}
    \label{fig:templates}

\end{figure}

In the onboarding stage, we render a set  of RGB synthetic templates and extract their visual descriptors using DINOv2~\cite{oquab2023dinov2}. 
To ensure robust object segmentation under different orientations, we render CAD models under 42 viewpoints as shown in Figure~\ref{fig:templates}. These 42 viewpoints are defined by the icosphere primitive of Blender~\footnote{\texttt{bpy.ops.mesh.primitive\_ico\_sphere\_add()}}  which has been shown in \cite{nguyen2022templates} to provide well-distributed view coverage of CAD models for robust template matching. Additionally, we experiment with denser viewpoints by dividing each triangle of the icosphere into four smaller triangles. The rendering process results in a total of $\NumTestObjects \NumViews$ templates, where $\NumTestObjects$ is the number of CAD models and $\NumViews$ is the number of viewpoints. 
We then crop the templates with the ground-truth bounding boxes and use the DINOv2 \texttt{cls} tokens as their visual descriptors $\ReferenceDescriptor$ of size $\NumTestObjects \times \NumViews \times \DescriptorSize$. By default, we use $\NumViews = 42$ and $\DescriptorSize = 1024$.

\subsection{Proposal stage}
\label{sec:proposal}

For each testing RGB image, we use SAM~\cite{kirillov2023segment} or FastSAM~\cite{zhao2023fast} with a default configuration to generate a set of $\NumProposals$ unlabeled proposals, where each proposal $i$ is defined by a mask $\Mask_i$. $\NumProposals$ is not fixed and varies depending on the content of the input RGB image.

To compute the visual descriptor for each proposal $i$, we first remove the background of the input image using the corresponding mask $\Mask_i$. Subsequently, we crop the image using the model bounding box derived from $\Mask_i$. Since each proposal mask $\Mask_i$ has a different bounding box size, parallel processing becomes unfeasible. To overcome this, we add a simple image processing step  including scaling and padding in order to resize all proposals to a consistent size of $224 \times 224$. This standardization enables efficient parallel processing of proposals in a single batch.
Then, we extract the DINOv2 \texttt{cls} tokens from the processed proposals and use them as their visual descriptors $\ProposalDescriptor$ of size $\NumProposals \times \DescriptorSize$.

\subsection{Matching stage}
\label{sec:matching}

The goal of the matching stage is to assign each proposal $i$ an object ID $\ObjectID_i$ and a confidence score $\Score_i$. To this end, we compare each proposal descriptor in $\ProposalDescriptor$ with each template descriptor in $\ReferenceDescriptor$ using the cosine similarity. This comparison step produces a similarity matrix of size $\NumProposals \times \NumTestObjects \times \NumViews$.

\customparagraph{View aggregation.} By aggregating the similarity scores over all $\NumViews$ templates for each CAD model, we obtain a matrix of size $\NumProposals \times \NumTestObjects$. This matrix represents the similarity between each proposal $\text{p}_i$ and each CAD model. We experiment with different aggregation functions, such as Mean, Max, Median, and Mean of top $k$ highest, noted $\text{Mean}_k$, and find that $\text{Mean}_k$ yields the best results.

\customparagraph{Object ID assignment.} To assign the object ID $\ObjectID_i$ and confidence score $\Score_i$ to each proposal, we simply apply the argmax and max functions on the similarity matrix $\NumProposals \times \NumTestObjects$ over the $\NumTestObjects$ objects. This yields a similarity matrix of size $\NumProposals$ defining the confidence score for $\NumProposals$ proposals.

\customparagraph{Output.} At the end of the matching stage, we obtain a set of labeled proposals, where each proposal is defined as $\{\Mask_i, \ObjectID_i, \Score_i\}$, where $\Mask_i$ is the modal mask (\ie, a mask covering the visible part of the object surface~\cite{sundermeyer2023bop}), $\ObjectID_i$ is the object ID, and $\Score_i$ is the confidence score. Some of these proposals may still be incorrectly labeled. To address this, it is possible to apply a threshold $\delta$ on the confidence score threshold. The figures in Section~\ref{sec:comparison} show CNOS's segmentation results with $\delta=0.5$.

\vspace{-0.2cm}
\section{Experiments}

\label{sec:experiments}
In this section, we describe the experimental setup~(Section~\ref{sec:exp_setup}), compare our method with previous works \cite{chen20233d, he2017mask, zebrapose} on the seven core datasets of the BOP challenge \cite{sundermeyer2023bop} (Section \ref{sec:comparison}), and conduct an ablation study focused on the accuracy under different aggregating functions and different numbers of rendering viewpoints, and on the run-time (Section \ref{sec:ablation}). Finally, we discuss the use of CNOS in a pipeline for 6D pose estimation of novel objects (Section \ref{sec:discussion}) or in CAD-free novel object segmentation.

\subsection{Experimental setup}
\label{sec:exp_setup}

\newcommand{\omitdetection}[1]{}
\definecolor{navyblue}{RGB}{191, 209, 229} 
\definecolor{light_yellow}{RGB}{255,243,194}

\begin{table*}[t]
  \centering
    
 \setlength{\tabcolsep}{2.5pt}
    \scalebox{0.90}{
  \begin{tabular}{lrl lcc ccccccc}
  
 \toprule  
 & & \multirow{3}{*}{Method} & \multirow{3}{*}{Rendering}  &  \multicolumn{8}{c}{BOP Datasets} &  \\ 
 
 \cmidrule(lr){5-11}
 & &  &  & \textsc{lm-o} & \textsc{t-less} & \textsc{tud-l} & \textsc{ic-bin} & \textsc{itodd} & \textsc{hb} & \textsc{ycb-v} & Mean \\

  \midrule  %
\parbox[t]{2mm}{\multirow{4}{*}{\rotatebox[origin=c]{90}{\small Supervised}}} & {\color{teal}\scriptsize 1} & Mask R-CNN~\cite{he2017mask} (Synth)  & - & 37.5 & 51.7& 30.6 &31.6& 12.2 &47.1& 42.9 & 36.2 \\ %
& {\color{teal}\scriptsize 2}  & Mask R-CNN~\cite{he2017mask} \hspace{0.5ex} (Real) & - &  37.5 & 54.4& 48.9 &31.6& 12.2 &47.1& 42.9 & 39.2\\ %
& {\color{teal}\scriptsize 3} & ZebraPose~\cite{zebrapose} \hspace{1.05ex} (Synth) & - & \cellcolor{navyblue} 50.6 & 62.9 & 51.4 & \cellcolor{navyblue}37.9 & \cellcolor{navyblue}36.1 & \cellcolor{navyblue}64.6 & 62.2 & 52.2\\ %
& {\color{teal}\scriptsize 4} & ZebraPose~\cite{zebrapose} \hspace{2.05ex} (Real)  & - &   \cellcolor{navyblue}50.6 & \cellcolor{navyblue}70.9 & \cellcolor{navyblue}70.7 & \cellcolor{navyblue}37.9 & \cellcolor{navyblue}36.1 & \cellcolor{navyblue}64.6 & \cellcolor{navyblue}74.0 & \cellcolor{navyblue}57.8\\ %

 \midrule %

\parbox[t]{2mm}{\multirow{4}{*}{\rotatebox[origin=c]{90}{\small Unsupervised}}} &  {\color{teal}\scriptsize 5} &  Chen~\textit{et al.}~\cite{chen20233d}   & - & 17.6 & 9.6 &24.1 & 18.7& 6.3 & 31.4 & 41.9 & 21.4\\ %
 
& {\color{teal}\scriptsize 6} & CNOS (SAM) & Pyrender\cite{Pyrender} & 33.3 & 38.3 & 35.8 & 27.2 & 14.8 & 45.9 & 57.6  & 36.1 \\ %

& {\color{teal}\scriptsize 7} &  CNOS (SAM) & BlenderProc \cite{denninger2019blenderproc} & 39.6 & \cellcolor{light_yellow} 39.7 & 39.1 & \cellcolor{light_yellow} 28.4 & \cellcolor{light_yellow} 28.2 & 48.0 & 59.5  & 40.4 \\ %

& {\color{teal}\scriptsize 8} & CNOS (FastSAM) & BlenderProc \cite{denninger2019blenderproc}  & \cellcolor{light_yellow} 39.7 & 37.4 & \cellcolor{light_yellow} 48.0 &  27.0 &  25.4 & \cellcolor{light_yellow} 51.1 & \cellcolor{light_yellow} 59.9  & \cellcolor{light_yellow} 41.2 \\ %




  \bottomrule
  \end{tabular}
    }
  \caption{{\bf Comparison of CNOS with \cite{he2017mask, zebrapose, chen20233d} on the seven core datasets of the BOP challenge~\cite{sundermeyer2023bop}.} Mask R-CNN and ZebraPose are retrained specifically on the target objects with renderings of the CAD models (noted as ``Synth") or real images of the object (noted as ``Real"). We classify these methods as ``supervised''. CNOS and Chen \emph{et. al.}~\cite{chen20233d} are classified as ``unsupervised'' as these methods require no retraining for novel objects. We report the AP metric (higher is better) using the protocol from \cite{sundermeyer2023bop}. The best supervised results are highlighted in \colorbox{navyblue}{blue} and the best unsupervised results in \colorbox{light_yellow}{yellow}. CNOS not only significantly outperforms \cite{chen20233d} under the same settings but also surpasses the supervised method Mask R-CNN, highlighting its ability to generalize.
  }
  \vspace{0.3cm}
  \label{tab:bop}
\end{table*}
\begin{figure*}[t!]
\newlength{\imagewidth}
\setlength\imagewidth{3.05cm}
\begin{center}
    \begingroup
    \setlength{\tabcolsep}{0.0pt}
    \renewcommand{\arraystretch}{1.0}
    \begin{tabular}{ c c c c c }
        \raisebox{0.35\imagewidth}{\makecell[b]{LMO\\ (8 objects)}}&\includegraphics[height=\imagewidth]{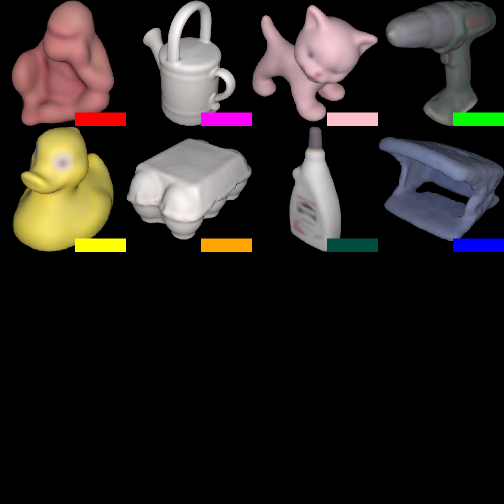} \hspace{0.25ex} &
        \includegraphics[height=\imagewidth]{figures/qualitative/lmo_000338.png} \hspace{0.25ex} &
        \includegraphics[height=\imagewidth]{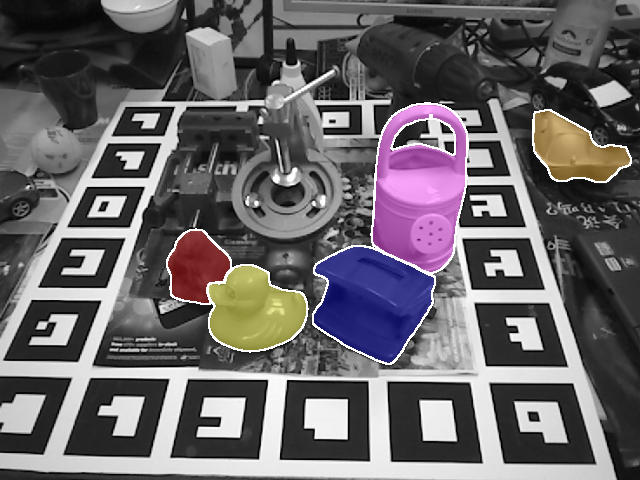} \hspace{0.25ex} &
        \includegraphics[height=\imagewidth]{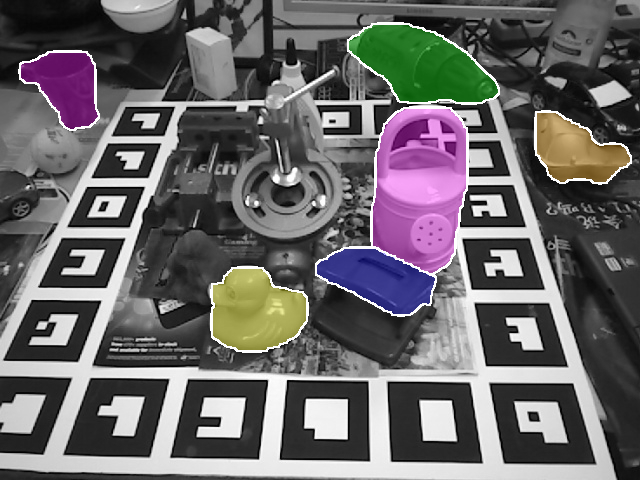} \\

        \raisebox{0.35\imagewidth}{\makecell[b]{HB\\ (33 objects)}} &\includegraphics[height=\imagewidth]{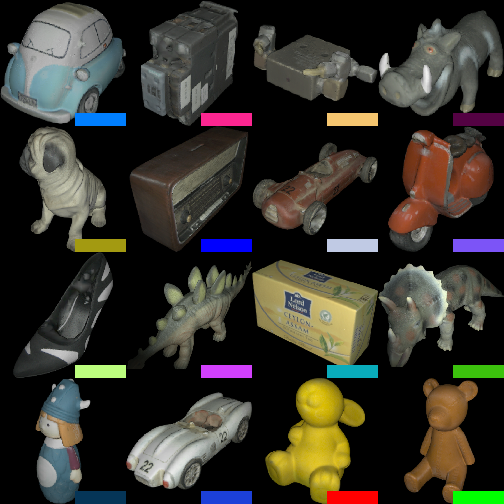} \hspace{0.25ex} &
        \includegraphics[height=\imagewidth]{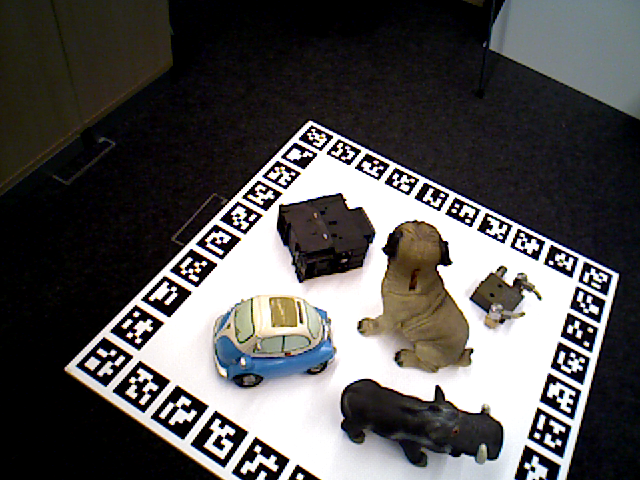} \hspace{0.25ex} &
        \includegraphics[height=\imagewidth]{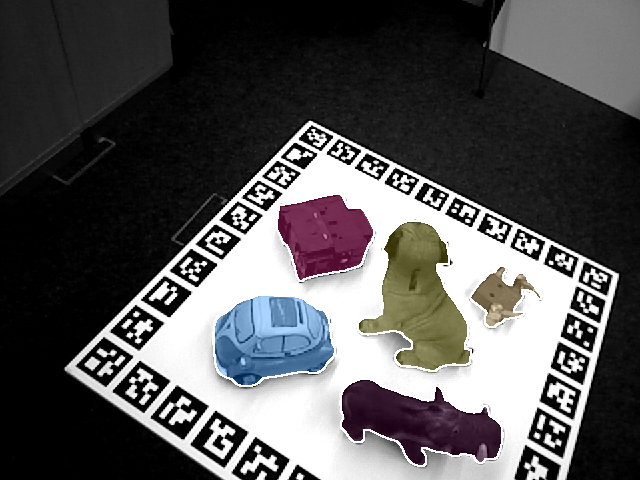} \hspace{0.25ex} &
        \includegraphics[height=\imagewidth]{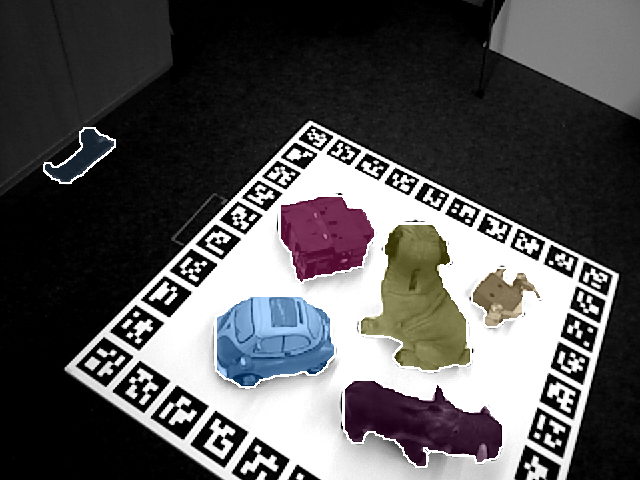} \\

        \raisebox{0.35\imagewidth}{\makecell[b]{YCB-V\\ (21 objects)}} &\includegraphics[height=\imagewidth]{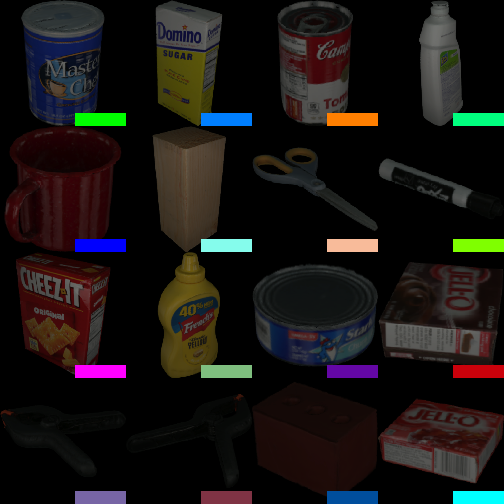} \hspace{0.25ex} &
        \includegraphics[height=\imagewidth]{figures/qualitative/ycbv_scene55_162.png} \hspace{0.25ex} &
        \includegraphics[height=\imagewidth]{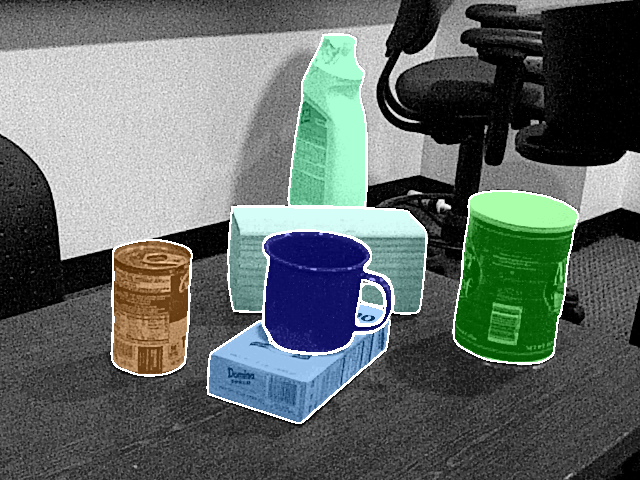} \hspace{0.25ex} &
        \includegraphics[height=\imagewidth]{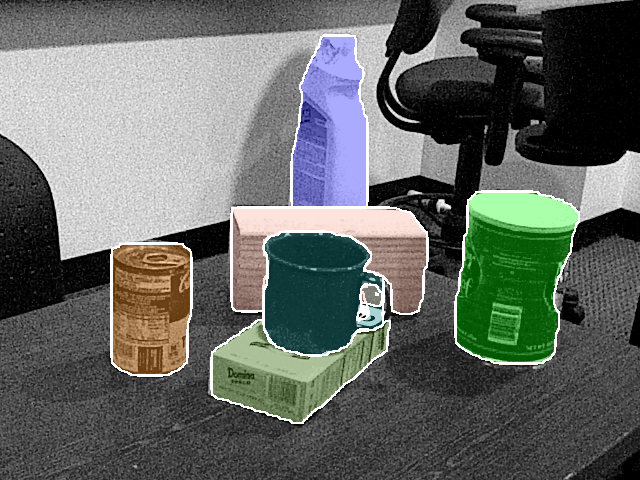}\\

        & Input CAD models & Input RGB & CNOS (SAM) & CNOS (FastSAM)\\
    \end{tabular}
    \endgroup
    \caption{
    \textbf{Qualitative results on LM-O ~\cite{brachmann-eccv14-learning6dobjectposeestimation}, HB~\cite{kaskman2019homebreweddb} and YCB-V~\cite{Xiang2018-dv}}. The first column shows the input CAD models. In cases where there are more than 16 models, we only show the first 16 to ensure better visibility. The second column show the input RGB image and the last two columns depict the detections produced by CNOS (SAM) and CNOS (FastSAM) with confidence scores greater than 0.5. \nguyen{Interestingly, in the last row, even though the segmentation proposals in CNOS (SAM) and CNOS (FastSAM) are very similar, their final labels differ for a few objects. This inconsistency arises from DINOv2-based classification of the proposals as discussed in Section \ref{sec:comparison}.} 
    }
    \label{fig:qualitative}
\end{center}
\end{figure*}

\paragraph{Datasets.} We evaluate our method on the test set of seven core datasets of the BOP challenge \cite{sundermeyer2023bop}: LineMod Occlusion (LM-O)~\cite{brachmann-eccv14-learning6dobjectposeestimation}, T-LESS~\cite{hodan-wacv17-tless}, TUD-L~\cite{Hodan_undated-sl}, IC-BIN~\cite{doumanoglou2016recovering}, ITODD~\cite{drost2017introducing}, HomebrewedDB (HB)~\cite{kaskman2019homebreweddb} and YCB-Video (YCB-V)~\cite{Xiang2018-dv}. In total, the datasets include 132 different objects shown in cluttered scenes with occlusions. The objects are of verious types: textured or untextured, symmetric or asymmetric, household or industrial.

\customparagraph{Evaluation metric.} We evaluate our method using the Average Precision (AP) metrics, following the COCO metric and the BOP challenge evaluation protocol \cite{sundermeyer2023bop}. The AP metric is calculated as the mean of AP values at different Intersection over Union (IoU) thresholds ranging from 0.50 to 0.95 with an increment of 0.05.

\customparagraph{Baselines.} We compare our method with Chen~\textit{et al.} \cite{chen20233d}, the most relevant work to ours. They use a three-stage CAD-based object segmentation approach, incorporating SAM \cite{kirillov2023segment} for image segmentation and ImageBlind \cite{girdhar2023imagebind} for visual descriptor extraction. Their use of 72 templates per CAD model resulted in the best performance according to their paper.  Additionally, we compare our method with two relevant supervised methods from the BOP challenge \cite{sundermeyer2023bop}: Mask R-CNN \cite{he2017mask}, which was trained on real or synthetic training images specific to each dataset and used in CosyPose \cite{labbe-eccv20-cosypose}, and ZebraPose \cite{zebrapose}, which is currently the state-of-the-art for this task in the BOP challenge.

\customparagraph{Implementation details.} For the proposal stage, we use the default ViT-H SAM \cite{kirillov2023segment} or the default FastSAM \cite{zhao2023fast}, which has demonstrated promising results in terms of run-time efficiency. For extracting visual descriptors, we use the default ViT-H model of DINOv2 \cite{oquab2023dinov2}.

To further evaluate the performance of our method, we conducted a comparison using two sets of templates. The first set of templates was generated using Pyrender \cite{Pyrender} from 42 pre-defined viewpoints. It is worthy to note that Pyrender computes the Direct Illumination and it is extremely fast, takes on average 0.026 second per image. The second set of templates comprised 42 realistic rendering templates selected from the available synthetic images of the PBR-BlenderProc4BOP training set provided in the BOP challenge. These realistic templates were specifically chosen to closely match the orientations of the 42 predefined viewpoints in the first set. \nguyen{Since the PBR-BlenderProc4BOP training images possibly have occlusions, we chose only images where the target objects are fully visible. The templates of target objects are finally obtained by making the background black using the ground-truth mask and cropping regions with the ground-truth bounding box.}


In order to maintain a consistent run-time across all datasets, we resize the images while preserving their aspect ratio. Specifically, we ensure that the width of each input RGB testing image is fixed at 640 pixels. All our experiments were conducted on a single V100 GPU.

\vspace{-0.2cm}
\subsection{Comparison with the state of the art}
\label{sec:comparison}


In Table~\ref{tab:bop}, we show that CNOS outperforms Chen~\textit{et al.}~\cite{chen20233d} by a significant margin of absolute 19.8\% AP. Furthermore, despite not being trained on the testing objects of the BOP datasets, our method surpasses the performance of Mask R-CNN \cite{he2017mask} used in CosyPose \cite{labbe-eccv20-cosypose}, which was specifically trained on these objects. This highlights the generalization capability of our method.

We qualitatively found that the generated segmentation proposals usually include ones that are very well aligned with the target object instances, and that most mistakes are due to erroneous DINOv2-based classification of the proposals. Improving the proposal classification would be crucial to close the gap between CNOS and supervised state-of-the-art approaches such as ZebraPose.

In Figure \ref{fig:qualitative}, we show qualitative results of our method on LM-O ~\cite{brachmann-eccv14-learning6dobjectposeestimation}, HB~\cite{kaskman2019homebreweddb} and YCB-V~\cite{Xiang2018-dv} datasets.

\subsection{Ablation study}
\label{sec:ablation}
\nguyen{\paragraph{Model size vs. run-time.} We present the results for FastSAM and DINOv2 using various base models in Table \ref{tab:ablation_model}, highlighting the trade-off between accuracy and run-time.}

\customparagraph{Rendering.} Table \ref{tab:bop} demonstrates the performance of our method using two types of rendering: Pyrender \cite{Pyrender} in row 6 and BlenderProc \cite{denninger2019blenderproc} in row 7. The results indicate that incorporating realistic rendering significantly reduces the domain gap between synthetic and real images, yielding a 4.3\% improvement in the AP metric.

\customparagraph{Number of viewpoints.} As shown in Table \ref{tab:viewpoint}, using more viewpoints does not bring any improvement compared to the coarse viewpoints. This can be explained by the fact that the current set of 42 coarse viewpoints already provides sufficient coverage of the 3D objects.

\begin{table}[t]
\centering
\setlength{\tabcolsep}{2.5pt}
\scalebox{0.87}{
\begin{tabular}{llcc}
\toprule  
  Segmentation model &  Descriptor model & AP & Run-time (s)\\ 
\midrule  %
 FastSAM-s &  ViT-s &  32.1 &  0.18 \\ %
 FastSAM-s &  ViT-l (default) &  33.8 &  0.25 \\
 FastSAM-x (default) &  ViT-s &  38.0 &  0.27 \\ %
 FastSAM-x (default) &  ViT-l (default) &  39.7 &  0.33 \\
\bottomrule
\end{tabular}
}
\caption{{\bf Ablation study of different FastSAM segmentation models~\cite{zhao2023fast} and DINOv2 descriptor models~\cite{oquab2023dinov2} on LM-O.} 
}
\label{tab:ablation_model}
\end{table}

\begin{table}[t]
\centering
\setlength{\tabcolsep}{2.5pt}
\scalebox{0.87}{
\begin{tabular}{l cc}

\toprule  
\multirow{3}{*}{Method} &  \multicolumn{2}{c}{Viewpoint density}  \\ 

\cmidrule(lr){2-3}
 &  Coarse (42) & Dense (162)  \\

\midrule  %
 CNOS (SAM) & \bf 39.6 &  39.5  \\ %
 CNOS (FastSAM) & \bf 39.7 &   39.7  \\ 

\bottomrule
\end{tabular}
}
\caption{{\bf Ablation study of the number of viewpoints on LM-O.} The denser viewpoints are created by subdividing each triangle of the icosphere (used to create coarse viewpoints) into four triangles.}
\label{tab:viewpoint}
\end{table}

\customparagraph{Aggregating function.} In Table \ref{tab:aggregating}, we explore different types of the function for aggregating the similarities between descriptors of templates and proposals. Among the tested functions, $\text{Mean}_k$ ($k$=5), which is the average of the $k$ highest similarity, achieves the best performance.

\begin{table}[t]
\centering
\setlength{\tabcolsep}{2.5pt}
\scalebox{0.9}{
\begin{tabular}{l cccc}
\toprule  
\multirow{2}{*}{Method} &  \multicolumn{4}{c}{Aggregating function}  \\ 

\cmidrule(lr){2-5}
&  Mean & Median & Max & $\text{Mean}_k$ \\
\midrule  %
 CNOS (SAM) & 36.6 &  34.9 &  39.1 &  \bf 39.6 \\ %
 CNOS (FastSAM) & 36.2 & 33.8 &  39.7 &  \bf 39.7 \\ 
 \midrule
Average & 36.40 & 34.40 &  39.40 &  \bf 39.65 \\
\bottomrule
\end{tabular}
}
\caption{{\bf Ablation study of aggregating functions on LM-O.} 
Using the $\text{Mean}_k$ function, which calculates the average of the top $k$ ($k=5$) highest values, yields the best accuracy.}
\label{tab:aggregating}
\end{table}

\customparagraph{Run-time.} In Table \ref{tab:runtime}, we present the average run-time of each stage in our method for a given CAD model. In the onboarding stage, the average rendering time for one image with Pyrender\cite{Pyrender} is 0.026 second while with BlenderProc~\cite{denninger2019blenderproc} is around 1 second per image on a single V100 GPU. It is important to note that the onboarding stage is performed once for each CAD model. In terms of run-time, the onboarding stage is clearly bottlenecked by the generation of templates, while the proposal stage is currently bottlenecked by the segmentation algorithm.

\begin{table}[t]
\centering
\setlength{\tabcolsep}{2.5pt}
\scalebox{0.85}{
\begin{tabular}{l ccc}

\toprule  
\multirow{3}{*}{Method} &  \multicolumn{3}{c}{Run-time (second)}  \\ 

\cmidrule(lr){2-4}
 &  Onboarding & Proposal & Matching \\

\midrule  %
 CNOS (SAM, Pyrender) & 1.22 & 1.58 &  0.13  \\ %
 CNOS (FastSAM, Pyrender) & 1.22 & 0.22 &  0.12 \\ 
CNOS (FastSAM, PBR) & 42.1 & 0.22 &  0.12 \\


\bottomrule
\end{tabular}
}
\caption{{\bf Run-time.} 
We report the run-time of each stage of CNOS on a single V100 GPU. The run-time of the onboarding stage includes both the rendering time and the visual descriptor extraction time for each CAD model.}
\label{tab:runtime}
\end{table}

\begin{figure}[!t]
\newlength{\imageheight}
\setlength\imageheight{1.3cm}
\centering
\setlength\lineskip{1.5pt}
\setlength\tabcolsep{1.5pt} 
{\small
\begin{tabular}{c}
\begin{tabular}{
>{\centering\arraybackslash}m{\imageheight}
>{\centering\arraybackslash}m{\imageheight}
>{\centering\arraybackslash}m{\imageheight}
>{\centering\arraybackslash}m{\imageheight}
>{\centering\arraybackslash}m{\imageheight}
>{\centering\arraybackslash}m{\imageheight}
}
Proposal &  Top 1 & Top 2 & Top 3 & Top 4 & Top 5\\ 
\includegraphics[height=\imageheight, ]{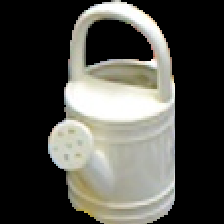}&
\includegraphics[height=\imageheight, ]{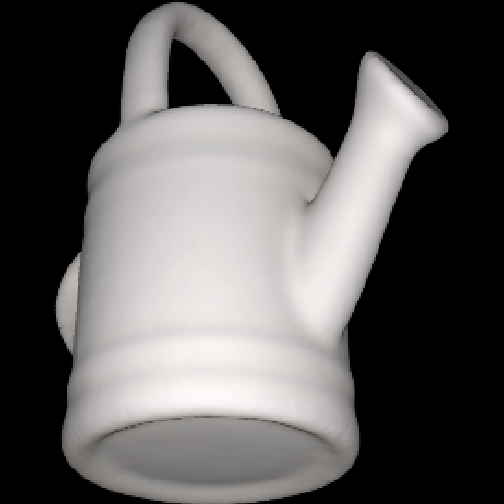}&
\includegraphics[height=\imageheight, ]{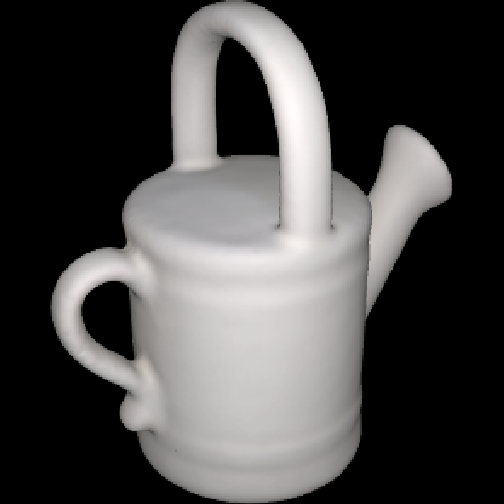}&
\includegraphics[height=\imageheight, ]
{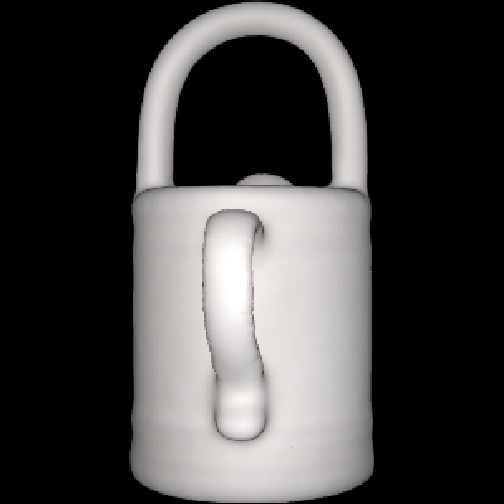}&
\includegraphics[height=\imageheight, ]{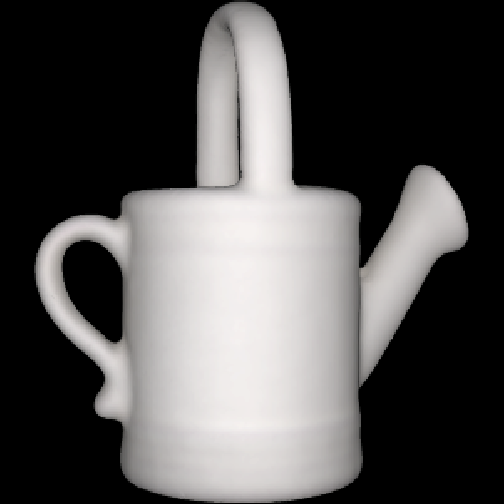}&
\includegraphics[height=\imageheight, ]{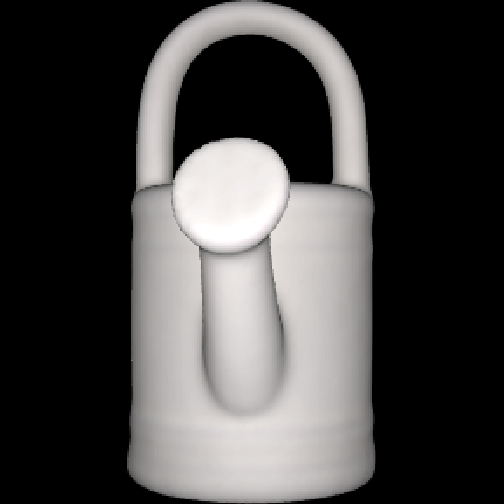}\\

\includegraphics[height=\imageheight, ]{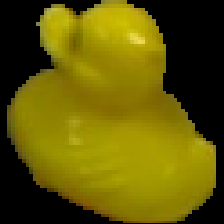}&
\includegraphics[height=\imageheight, ]{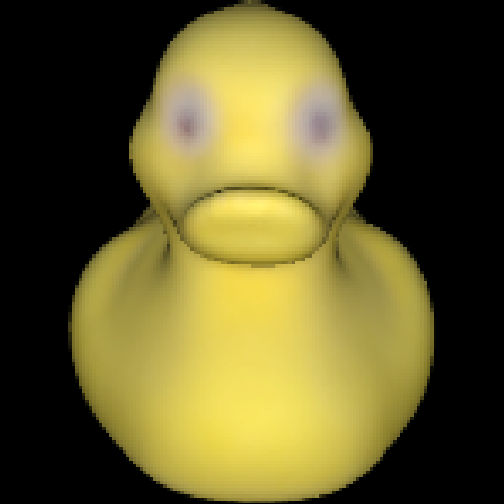}&
\includegraphics[height=\imageheight, ]{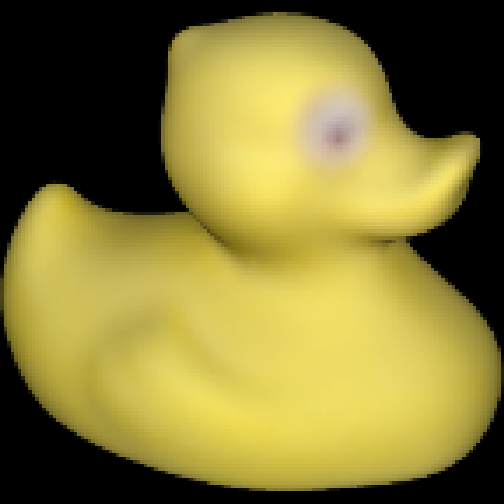}&
\includegraphics[height=\imageheight, ]{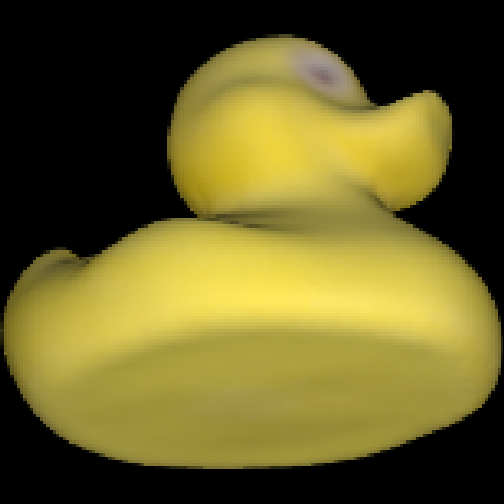}&
\includegraphics[height=\imageheight, ]{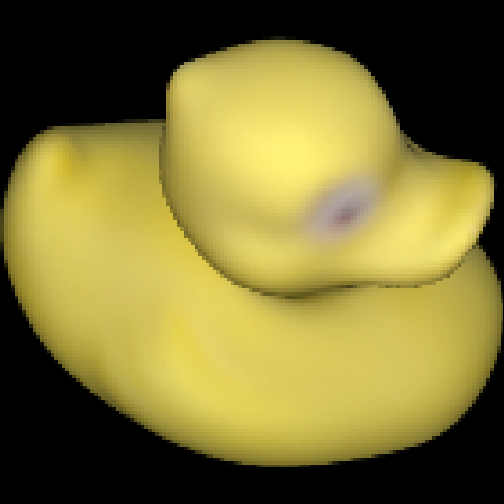}&
\includegraphics[height=\imageheight, ]{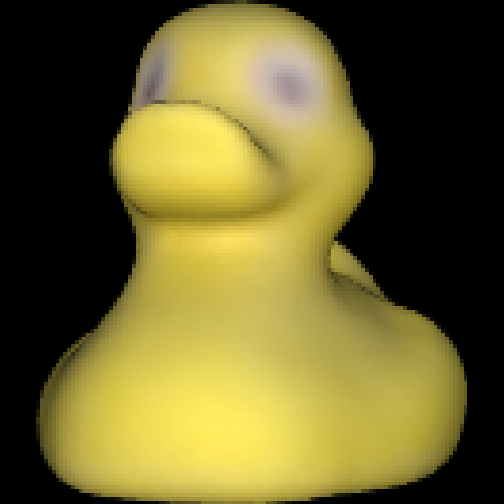}\\

\includegraphics[height=\imageheight, ]{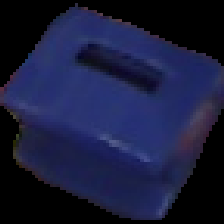}&
\includegraphics[height=\imageheight, ]{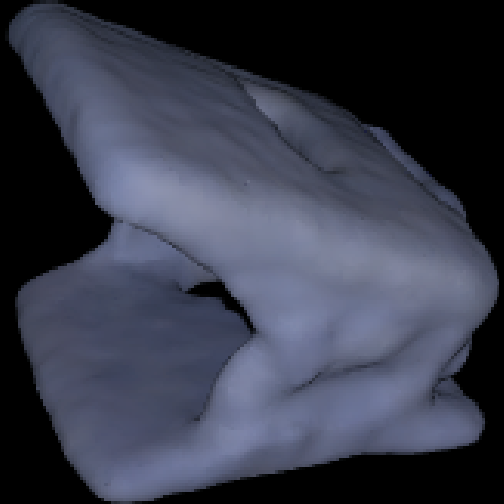}&
\includegraphics[height=\imageheight, ]{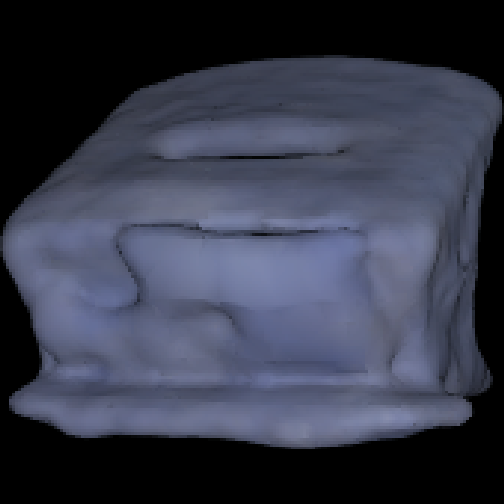}&
\includegraphics[height=\imageheight, ]{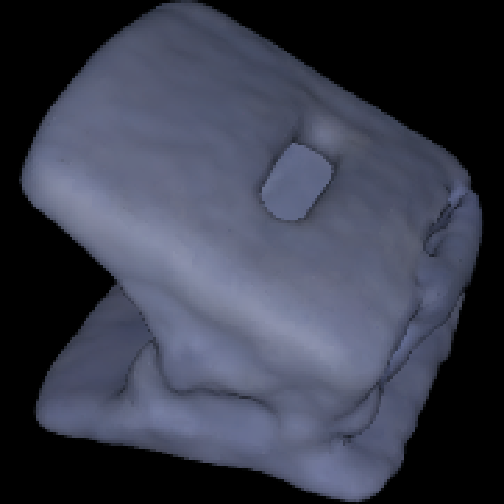}&
\includegraphics[height=\imageheight, ]{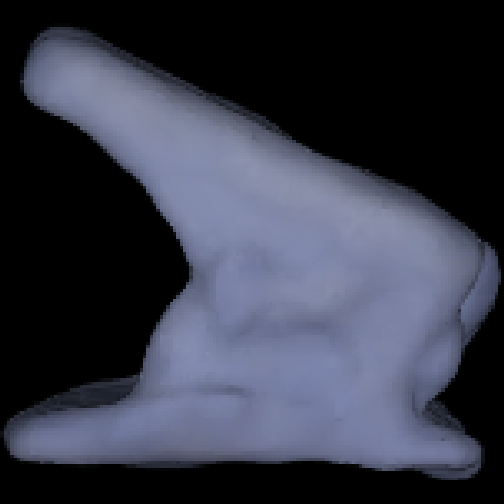}&
\includegraphics[height=\imageheight, ]{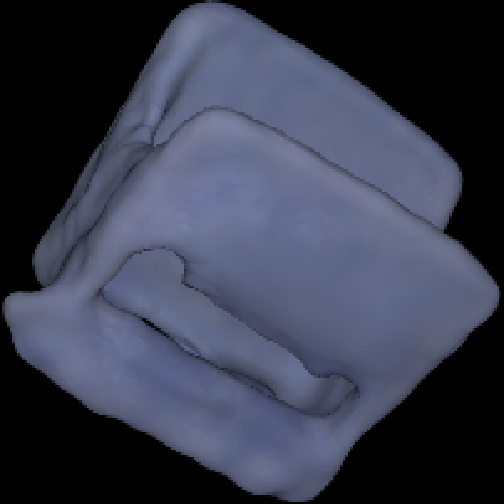}\\

\end{tabular}\\
\end{tabular}
}
\vspace{-0.2cm}
\caption{{\bf Visualization of the nearest neighbors.} We show proposals along with five retrieved templates with the most similar DINOv2 \texttt{cls} tokens. The retrieved templates correspond to the same object but to poses that do not match the pose in the proposals -- this suggests that the DINOv2 \texttt{cls} token can be effectively used to recognize the objects, but not to estimate the pose. 
}
\label{fig:neighbor}
\end{figure}

\subsection{Discussion}
\label{sec:discussion}
\customparagraph{Pose intialization.} \tom{Our intention was originally to use the DINOv2 \texttt{cls} token not only to recognize the object but also to estimate its initial pose that could be refined in a subsequent step.
However, as illustrated in Figure \ref{fig:neighbor}, this approach did not yield successful results, as the DINOv2 \texttt{cls} token seems to carry sufficient information about the object identity but not about the object pose.}

\customparagraph{CAD-free novel object segmentation.} \nguyen{In this work, we focus on CAD-based novel object segmentation. However, the proposed CNOS method could be seamlessly adapted to address one-shot or few-shot novel object segmentation settings, where only one or a few reference images are available and CAD models are unknown. Specifically, the reference descriptors could be extracted directly from the available reference image(s), while the rest of the pipeline could be kept untouched.}

\section{Conclusion}
We presented a simple yet powerful method for novel object segmentation solely based on their CAD models, without the need of any training. The method achieves a surprisingly high accuracy, comparable to previous supervised methods trained on large-scale annotated datasets. We hope that CNOS will serve as a standard baseline for CAD-based novel object segmentation and will be employed as the initial stage of novel object pose estimation pipelines.

\vspace{0.2cm}
{\small \noindent\textbf{Acknowledgments.} We thank Nermin Samet for helpful discussions. This research was produced within the framework of Energy4Climate Interdisciplinary Center (E4C) of IP Paris and Ecole des Ponts ParisTech, and was supported by 3rd \emph{Programme d’Investissements d’Avenir} [ANR-18-EUR-0006-02] and by the Foundation of Ecole polytechnique (Chaire “Défis Technologiques pour une Énergie Responsable” financed by TotalEnergies). This work was performed using HPC resources from GENCI–IDRIS 2022-AD011012294R2.}

{\small
\bibliographystyle{ieee_fullname}
\bibliography{cleaned_refs}
}

\end{document}